\documentclass[10pt,twocolumn,letterpaper]{article}

\usepackage{cvpr}
\usepackage{times}
\usepackage{epsfig}
\usepackage{graphicx}
\usepackage{amsmath}
\usepackage{amssymb}
\usepackage{booktabs}
\usepackage{authblk}
\usepackage{soul}

\usepackage[pagebackref=true,breaklinks=true,letterpaper=true,colorlinks,bookmarks=false]{hyperref}

\cvprfinalcopy 

\def\cvprPaperID{4729} 

\ifcvprfinal\fi
\begin{document}

\title{An Attempt towards Interpretable Audio-Visual Video Captioning}

\author[1]{Yapeng Tian}
\author[1]{Chenxiao Guan}
\author[2]{Justin Goodman}
\author[3]{Marc Moore}
\author[1]{Chenliang Xu}
\affil[1]{Department of Computer Science, University of Rochester}
\affil[2]{Department of Computer Science, University of Maryland}
\affil[3]{Department of Computer Science, Mississippi State University}

\maketitle

\begin{abstract}
Automatically generating a natural language sentence to describe the content of an input video is a very challenging problem. It is an essential multimodal task in which auditory and visual contents are equally important. Although audio information has been exploited to improve video captioning in previous works, it is usually regarded as an additional feature fed into a black box fusion machine. How are the words in the generated sentences associated with the auditory and visual modalities? The problem is still not investigated. In this paper, we make the first attempt to design an interpretable audio-visual video captioning network to discover the association between words in sentences and audio-visual sequences. To achieve this, we propose a multimodal convolutional neural network-based audio-visual video captioning framework and introduce a modality-aware module for exploring modality selection during sentence generation. Besides, we collect new audio captioning and visual captioning datasets for further exploring the interactions between auditory and visual modalities for high-level video understanding. Extensive experiments demonstrate that the modality-aware module makes our model interpretable on modality selection during sentence generation. Even with the added interpretability, our video captioning network can still achieve comparable performance with recent state-of-the-art methods. 
\end{abstract}

\section{Introduction}

\label{intro}
\begin{figure}[t]
\centering
\includegraphics[width=0.45\textwidth]{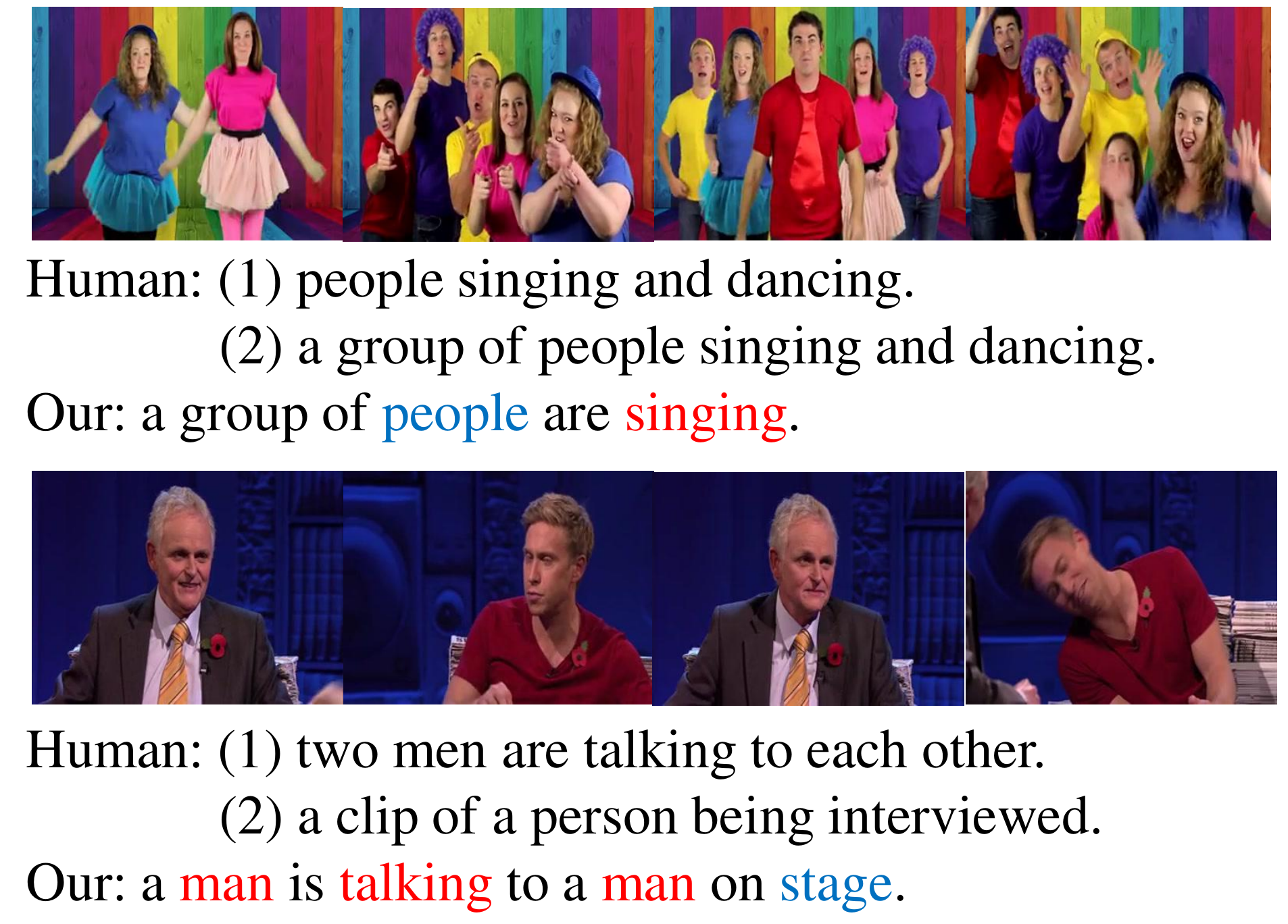}
\caption{Audio-visual video captioning with interpretability on modality selection during word generation. The automatically detected audio activated words and visual activated words are highlighted with \textbf{red} and \textbf{blue} texts, respectively. In the first example, visual modality is dominated for generating the \textit{people} and audio content is more informative for predicting the \textit{singing}. }
\label{demo}
\end{figure}

Video captioning aims to automatically generate a natural language sentence to describe an input video, where the community focuses on captioning unconstrained web videos. These videos usually contain complex spatiotemporal-dynamic scenes and rich visual contents that make the captioning task very challenging. Moreover, videos contain audio tracks that often reveal important in-scene and/or out-of-scene information; modeling them is essential towards comprehensive human-level understanding of videos. For the first example in Fig.~\ref{demo}, it is very difficult to recognize the singing event by only watching the video without sound, however, the audio content can clearly detect the singing event. Therefore, video captioning is indeed a multimodal problem in which both auditory and visual modalities play important roles. 

On one hand, the auditory modality has been exploited to improve video captioning performance in recent works, \eg, \cite{jin2016describing, ramanishka2016multimodal, shen2017weakly, xu2017learning}. These approaches usually fuse the audio features with features from other modalities, and then feed the encoded multimodal features into a decoder RNN, which exploits the fused features in a blind manner. Indeed, the commonly used RNN-based sequence-to-sequence architecture has inherent difficulties to perform modality-interpretable video captioning. When generating a word, beside using current provided/attended features and previous words, these models always exploit hidden states of the decoder RNN. The latter contain memorized information from different modalities, which make the models impossible to disentangle the contributions from individual modalities for predicting the word.

On the other hand, we have witnessed noticeable advances in learning from audio-visual data over the recent years. Many interesting problems have been studied, such as representation learning with cross-modal supervisions~\cite{aytar2016soundnet,owens2016ambient}, lip reading~\cite{chung2017lip}, and sound source separation~\cite{Gao_2018_ECCV}. More recently, audio-visual event localization tasks~\cite{Tian_2018_ECCV} are explored as a proxy to investigate the interaction between the two modalities. Despite the rich development, most works are focused on either low-level tasks, e.g., localization, and association, or learning a \textit{good} representation, the interplay of the two modalities in high-level captioning task has yet to be quantified. 

In this paper, we aim to disentangle such interplay of the two modalities and make the first attempt to interpretable audio-visual video captioning. Concretely, we propose a novel multimodal convolutional neural network (see Sec.~\ref{2DCNN})-based audio-visual video captioning framework without a RNN decoder to ease the design of interpretable structure, and introduce a modality-aware feature aggregation module (see Sec.~\ref{MAA}) with defined activation energy to distinguish which modality is more informative for generating words. With extensive experiments, we find that the proposed modality-aware module makes our network have interpretability on modality selection during word generation (two examples are illustrated in Fig.~\ref{demo}) and even with the added interpretability, our video captioning network can achieve comparable performance with recent state-of-the-art methods.

For further exploring the interactions between audio and visual modalities, we collect new audio captioning and visual captioning datasets and utilize audio, visual, and cross-modal captioning tasks as a proxy. The experiments validate that there is still a strong correlation between auditory and visual modalities even with the high-level video understanding tasks as the testbed, enabling cross-modal semantic inferring be possible for the future work.

Our paper makes the following contributions: (1) we propose a novel multimodal convolutional neural network-based audio-visual video captioning framework, which is friendly to design interpretable modality-aware structures; (2) based on the audio-visual captioning framework, we propose a novel modality-aware aggregation module with the defined activation energy to make the captioning model be more interpretable; (3) we collect two new datasets to explore audio-visual interactions for high-level video understanding. The experiments demonstrate that strong correlation between the auditory and visual modalities on captioning tasks enabling auditory modality to infer high-level semantics of the visual modality, and vice versa.  Dataset, code, and pre-trained models will be released. A video demo is available on \textcolor{magenta}{\url{https://www.youtube.com/watch?v=HC7Sn4-7iw0}}.

The rest of the paper is organized as follows: Sec.~\ref{related} introduces the related work. The proposed audio-visual video captioning network is described in Sec.\ref{net}. Experimental results are shown in Sec.~\ref{exp}. Sec.~\ref{conclude} gives the conclusion.

\section{Related Work}
\label{related}

In this Section, we introduce some related works on image captioning, video captioning, and vision-and-sound modeling.

\subsection{Image Captioning} 
\label{im_cap}
The dominant image captioning networks are based on the encoder-decoder framework~\cite{bahdanau2014neural}. Kiros~\etal~\cite{kiros2014multimodal} developed feed forward neural network-based multimodal neural language models to generate image descriptions. Recurrent neural networks (RNNs) were introduced in~\cite{chen2015mind,mao2014deep} to enhance the capacity of neural language models. Vinyals~\etal~\cite{vinyals2015show} proposed an end-to-end captioning network with a CNN encoder and LSTM-based sentence generator. Visual attention mechanisms~\cite{xu2015show,lu2017knowing,anderson2018bottom} were explored in image captioning networks to replace fixed CNN encoders and further improve captioning performance.

\begin{figure*}[t]
\centering
\includegraphics[width=\textwidth]{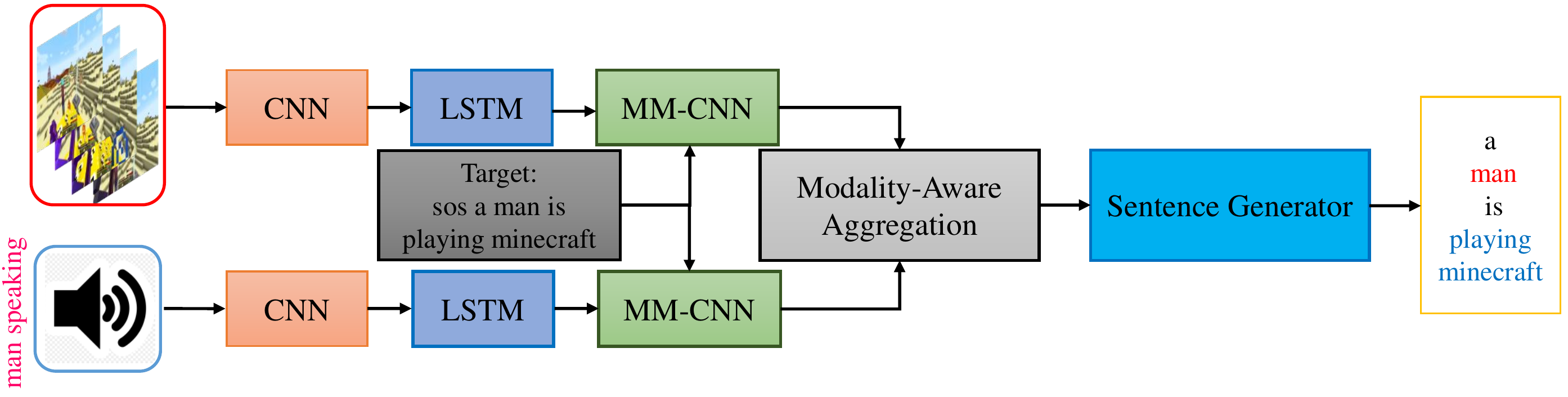}
\caption{The proposed MM-CNN-based audio-visual video captioning framework. During testing, words in the sentence will be predicted one-by-one. The input video frames only contain content of the video game, but there is man speaking sound in the audio channel. The word \textit{man} will be inferred from activated the auditory modality, and the words \textit{playing} and \textit{minecraft} are mainly from visual modality. We make modality selection decision based on values of audio activation energy and visual activation energy.}
\label{mm_cnn}
\end{figure*}

\subsection{Video Captioning}
\label{v_cap}
There are temporal dynamic scenes, rich visual contents, and auditory modality in videos. Therefore, video captioning is much more challenging than the image captioning task. Inspired by the early image captioning networks, Venugopalan~\cite{venugopalan2014translating} utilize a CNN encoder to extract visual features for sampled video frames and perform mean pooling over these features for video captioning. To explore temporal structures of videos, Venugopalan~\cite{venugopalan2015sequence} propose a sequence-to-sequence video captioning network, which exploits LSTMs to encode visual features from different video frames. Yao \etal utilize a spatial temporal 3D CNN and a temporal attention mechanism to explore local and global temporal structures in videos. Pan~\etal~\cite{pan2016hierarchical} propose a hierarchical recurrent neural encoder for exploiting temporal information. Zhang~\etal~\cite{Zheng_2017_CVPR} develop a dynamic fusion method to combine appearance and motion features for video captioning. Reinforcement learning is exploited in \cite{wang2018video} and \cite{chen2018less}.

Excepting the visual content, auditory modality and video tags have also been demonstrated that it can help to improve video captioning performance in~\cite{jin2016describing,ramanishka2016multimodal,chen2017video,xu2017learning,hori2017attention,shen2017weakly,wang2018watch}. Naive fusion (\eg concatenation in~\cite{ramanishka2016multimodal}) and attention fusion (\eg~\cite{hori2017attention}) are utilized in these methods to aggregate features from different modalities. Although these methods exploit the audio information, they fail to completely analyze the two modalities for video captioning due to essential difficulties existing in their RNN decoder-based frameworks. 

Unlike previous methods, in this paper, we focus on analyzing the auditory and visual modalities for audio-visual video captioning. To explore the associations between words in sentences and individual modalities, we introduce a multimodal convolution neural network-based interpretable video captioning framework with a modality selection-aware module.  

\subsection{Vison-and-Sound Modeling:}
\label{vsm}
Recently, research topics about vision and sound have attracted a lot of attentions. Aytar \etal~\cite{aytar2016soundnet} propose a SoundNet to learn audio features using a visual teacher network from massive unlabeled videos. Owens~\etal~\cite{owens2016ambient} adopt ambient sounds as a supervisory signal for learning visual models. To learn both a joint embeding for audio and visual content, Arandjelovic and Zisserman introduce a audio-visual corresponding prediction task. To learn audio and spatio-temporal visual representions, Owens and Efros~\cite{Owens_2018_ECCV}, and Korbar~\etal~\cite{korbar2018co} utilize the audio-visual temporal synchronization task as a proxy. Excepting representation learning, some works on sound source separation~\cite{Zhao_2018_ECCV,Gao_2018_ECCV}, sound source localization~\cite{senocak2018learning,Arandjelovic_2018_ECCV}, and audio-visual event localization~\cite{Tian_2018_ECCV} have also been studied. Unlike the previous works, in this paper, we will investigate audio-visual video captioning task and explore whether the audio modality can infer high-level semantics of visual modality and vice versa with the captioning problem as a proxy.

\section{Proposed Network}
\label{net}

First, we present the overall audio-visual video captioning framework in Sec.~\ref{framework}. Upon this framework, we propose our multimodal convolutional neural network in Sec.~\ref{2DCNN} and modality-aware aggregation net in Sec.~\ref{MAA}. Finally, we introduce the sentence generator in Sec.~\ref{SG}.

\subsection{Audio-Visual Video Captioning Network}
\label{framework}

Given an input visual and audio clip pair $\{V, A\}$, our audio-visual video captioning network aims to generate a natural language sentence $S = (s_1, s_2,\dots,s_{T_s})$ containing $T_s$ words. Unlike previous RNN-based encoder-decoder video captioning networks, in this paper, inspired by~\cite{elbayad2018pervasive}, we propose a 2D multimodal convolutional neural network-based audio-visual video captioning framework illustrated in Fig. \ref{mm_cnn}, which is capable of learning deep feature hierarchies and is more convenient for designing interpretable modules. The network mainly consists of five modules: feature extraction, temporal modeling, \textbf{m}ulti\textbf{m}odal \textbf{c}onvolutional \textbf{n}eural \textbf{n}etwork (MM-CNN), modality-aware aggregation module, and sentence generator. 

The feature extraction module utilizes pre-trained CNN models to extract visual features $v\in\mathcal{R}^{T_v\times D_v}$ and audio features $a\in\mathcal{R}^{T_a\times D_a}$ from the input visual clip $V$ and audio clip $A$. Here, we sample $T_v$ video frames from the given visual clip $V$ and $T_a$ seconds for the given audio clip $A$. Visual feature dimension for each frame is $D_v$ and audio feature dimension for each second audio segment is $D_a$. 

We observe that audio sequences, visual sequences and textual sequences (captioning sentences) even from same videos may have different temporal patterns. To explore temporal structures in each input modality, we use two separate LSTMs as the temporal modeling modules, which take visual feature $v$ and audio feature $a$ as inputs respectively. For a feature vector $m_t$ at the time step $t$, a LSTM will
update a hidden state $h_t$ and a memory cell state $c_t$:
\begin{align}
h_t, c_t = LSTM(m_t, h_{t-1}, c_{t-1}) \enspace,
\label{lstm}
\end{align}
where $m_t$ refers to the visual feature vector $v_t$ in $v$ or the audio feature vector $a_t$ in $a$ in our temporal modeling module. Here, the subscript $t$ indexes over $T_v$ video frames and $T_a$ audio segments, when we encoding the features over each modality, respectively. The module will perform temporal dependency modeling for audio and visual modality, respectively, and make certain implicit alignments with textual modality for the two modalities. 

Our language model consists of two MM-CNNs. Taking the aggregated hidden states $h^a\in\mathcal{R}^{T_a\times D_a}$ from the audio LSTM and the sentence $S$ as inputs, our audio-text MM-CNN will predict a joint deep audio-text embedding $F^{a}$. Similarly, we can predict a joint deep visual-text embedding $F^v$ from a visual-text MM-CNN with $h^v\in\mathcal{R}^{T_v\times D_v}$ and $S$ as inputs. We describe the detail of this process in Sec.~\ref{2DCNN}.

To deal with features extracted from the two modalities, we propose a modality-aware aggregation module (see Sec.~\ref{MAA}), which will perform modality selection and make our video captioning network interpretable in terms of quantifying the contributions from auditory and visual modalities, respectively. With the aggregated features, the sentence generator (see Sec.~\ref{SG}) will predict words parallel during training and one-by-one during inference. 


\subsection{Multimodal Convolutional Neural Network}
\label{2DCNN}
To predict deep joint embeddings for further sentence generation from audio hidden states $h^a$, visual hidden states $h^v$, and generated previous words in the $S$, we propose an autoregressive language model: MM-CNN. For individually handling auditory and the visual modalities, a visual-text MM-CNN and an audio-text MM-CNN are utilized to learn joint visual-text embedding and joint audio-text embedding, respectively. With the visual-text MM-CNN as an example, we introduce details about the model. 

The visual-text MM-CNN mainly contains two parts: visual-text tensor construction and joint deep visual-text feature extraction. 

\vspace{2mm}
\noindent\textbf{Tensor Construction:} \quad 
For a target sentence $S$, we first extract word embedding $e_{t}\in\mathcal{R}^{D_s}$ for each word $s_t$ in $S$ and then combine all words into a matrix $e\in\mathcal{R}^{T_s\times D_s}$. Given the aggregated visual hidden states $h^v\in\mathcal{R}^{T_v\times D_v}$ for a video clip $V$ and word embedding $e$ for the sentence $S$, we construct a 3D tensor $I^v\in\mathcal{R}^{T_s\times T_v\times D_{vs}}$,  
where $D_{vs} = D_v + D_s$ and $I_{ij}^{v} = [e_{i}$  $h^v_{j}]$.  Note that, for designing an autoregressive language model, the first word in the sentence $S$ will be set as $<sos>$. The visual-text tensor is the input of the joint deep feature learning module, which we will describe next.

\begin{figure}[t]
\centering
\includegraphics[width=8cm]{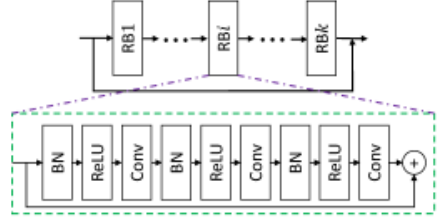}
\caption{Residual Network and residual blocks in the proposed MM-CNN.}
\label{fig:res}
\vspace{-5mm}
\end{figure}

\vspace{2mm}
\noindent\textbf{Joint Deep Feature Learning:} \quad 
To learn joint deep representations for visual and textual modalities, we feed the tensor $I^v$ into a deep residual 2D CNN network $f_v$. The joint visual-text embedding $F^v\in\mathcal{R}^{T_s\times T_v\times D_{vs}}$ can be obtain:
\begin{align}
F^v = f_v(I^v) \enspace.
\label{emb}
\end{align}
Following the deign of residual blocks in the ResNet and considering computation efficiency, we utilize the residual block layout as illustrated in Fig.~\ref{fig:res}. It consists of three convolutional layers, and the first two layers are $1\times1$ convolutional layers and the last one is a $3\times3$ convolutional layer. The output channels for the three layers are $D_{vs}\times2$, $\lfloor D_{vs}/2\rfloor$, and $D_{vs}$, respectively. To reduce computation complexity, we employ the $3\times3$ convolution operation on an input with smaller channel number by using the second $1\times1$ bottleneck layer. To preserve the capacity of the residual block, we use the first layer to expand the input feature dimension. Batch normalization~\cite{ioffe2015batch} is also adopted to stabilize network training and ReLU~\cite{nair2010rectified} is used as the nonlinear activation function. 

For building a language model, we need to make the network be autoregressive. Due to the sentence shifting operation, $1\times1$ convolution is essential autoregressive in our network\footnote{We shift the sentence one position and set the $<sos>$ at the first position. For predicting the real first word $S_2$ in the sentence, information from the input video clip and the embedding of $<sos>$ will be utilized.}. However, the $3\times3$ convolution will utilize future textual information. To address the issue, we exploit a masked $3\times3$ convolution as~\cite{elbayad2018pervasive}, which sets weights of convolution kernels at future textual positions be $0$. 

We stack $k$ residual blocks for learning high-level deep features and introduce a dense skip connection between the input features and the output features of the deep residual network for integrating low-level and high-level features. 

The proposed MM-CNN is capable of learning deep multimodal embeddings and the 2D convolution makes the network have the potential to well explore local multimodal sequence patterns. It can be easily extende to other multimodal tasks such as video question answering~\cite{zhu2017uncovering} and video-to-text retreival~\cite{mithun2017cmu}. 

Similarly, we can build a audio-text MM-CNN to predict the joint deep embedding $F^a\in\mathcal{R}^{T_s\times T_a\times D_{as}}$. Next, we will aggregate features over different time steps within the same modality and cross differet modalities for the sentence generation.

\subsection{Modality-Aware Aggregation}
\label{MAA}
The modality-aware aggregation module will adaptively select features over different time steps and cross different modalities for captioning generation. 

Given $F^a\in\mathcal{R}^{T_s\times T_a\times D_{as}}$ and $F^v\in\mathcal{R}^{T_s\times T_v\times D_{vs}}$, we first use two fully connected layers to align the two tensors with a same feature dimension $D_c$, and then construct a new tensor $F^c \in\mathcal{R}^{T_s\times T_c\times D_c}$ by concatenating the two tensors along the audio-visual channel, where $T_c = T_v + T_a$. 

Let $F_i^c\in\mathcal{R}^{T_c\times D_c}$ be the i-$th$ row of the $F^c$. We will use the $F_i^c$ to generate a feature vector $x_i\in\mathcal{R}^{D_c}$ and then predict the $(i+1)$-$th$ word $S_{i+1}$ with the $x_i$. The naive and simple way to generate the $x_i$ from $F_i^c$ by max-pooling or mean-pooling. However, the two methods are modality-ambiguous, which makes the modality selection be uninterpretable. Motivated from the Squeeze-and-Excitation design in~\cite{Hu_2018_CVPR}, we introduce a modality-aware aggregation module to compute $x_i$ from $F_i^c$:
\begin{align}
x_i = \sum_{j=1}^{T_c}w_{j}F_{ij}^c \enspace,
\label{se}
\end{align}
where the $F_{ij}^c\in\mathcal{R}^{D_c}$ and $w_j\in[0,1]$. We define activation energies for auditory and visual modality for measuring which modality is the dominant one that generates a noun or a verb word. The visual activation energy is defined as:
\begin{align}
e_i^v = \sum_{j=1}^{T_v}w_{j}^2 \enspace.
\label{v_energy}
\end{align}
Similarly, the audio activation energy can be computed as:
\begin{align}
e_i^a = \sum_{j=T_v+1}^{T_c}w_{j}^2 \enspace.
\label{v_energy}
\end{align}
When the generated word is a noun or a verb, if $e_i^v > e_i^a$, visual content is more important for generating the word; if $e_i^v < e_i^a$, the word is more related to the auditory modality. In this way, our model will have interpretable ability for modality selection during word generation.
The weights can be computed by:
\begin{align}
w_1,...,w_{T_c} = softmax(u) \enspace, \\
u = fc_3(\delta(fc_2(fc_1(F_i)))) \enspace, 
\end{align}
where the first Fully-Connected (FC) layer $fc_1$ aggregates features at each position $j$ of the $F_i$, $fc_1(F_i)\in\mathcal{R}^{T_c}$, the second and third FC layers corresponding to Squeeze and Excitation operations~\cite{Hu_2018_CVPR} having $\lfloor T_c/2\rfloor$ and $T_c$ output neurons, respectively. Here, $u\in\mathcal{R}^{T_c}$, and the $\delta$ is the ReLU activation function.

\subsection{Sentence Generator}
\label{SG}
 To predict the next word $S_{i+1}$, we compute its probability distribution over previous words $S_{1:i}$, the input visual clip $V$, and the input audio clip $A$:
\begin{align}
p(S_{i+1}|S_{1:i},V,A)=p(S_{i+1}|x_i)=softmax(Wx_{i}),
\label{prob.}
\end{align}
where the $W$ is a projection matrix and $x_i$ summarizes information from the input audio clip $A$, input visual clip $V$, and words before $i+1$. With network parameter $\theta$, it can be optimized by the cross entropy loss:
\begin{equation}
\mathcal{L(\theta)} = -\sum_{i=1}^{T_s}log(S_{i+1}|S_{1:i},V,A;\theta) \enspace.
\label{xe}
\end{equation}
There is no recurrent structures in the MM-CNN, the modality-aware aggregation, or the sentence generator. Therefore, our network can predict the whole sentence simultaneously during training. During inference, words will be generated one by one. 

\section{Experiments}
\label{exp}
First, we introduce datasets in Sec.~\ref{data}, the used visual, audio, and word representations in Sec.~\ref{feas}, evaluation metrics in Sec.~\ref{metric}, and implementation details in Sec.~\ref{ID}. Then, we compare the proposed video captioning network with the state-of-the-art methods in Sec.~\ref{sota}. Moreover, we show the results of the modality selection in Sec.~\ref{res}. Furthermore, we provide audio captioning, visual captioning and cross-modal captioning results in Sec.~\ref{cmc}. Finally, we discuss the effect of the proposed MM-CNN with different numbers of the residual block in Sec.~\ref{block}. 

\subsection{Datasets}
\label{data}
In this work, we train and evaluate the proposed audio-visual video captioning model on the MSR-VTT~\cite{xu2016msr}. MSR-VTT is a large-scale video Description dataset for bridging video and language, which has 10,000 video clips over 20 video categories with diverse video content and descriptions, as well as multimodal audio and video streams. Each clip is annotated with 20 natural language descriptions by AMT workers. Following the original split in MSR-VTT, we split the data according to 65\%:30\%:5\%, which corresponds to 6,513, 2,990 and 497 clips in the training, testing and validation sets, respectively.

Although the MSR-VTT~\cite{xu2016msr} contains audio-visual captioning annotations, it has no individual video captioning and audio captioning sentences which are desired for us to further explore correlation between the two modalities on the high-level captioning task. Therefore, we collected audio captioning and visual captioning annotations for a subset of the MSR-VTT, and introduce an audio captioning dataset (referred as AC) and a video captioning dataset (referred as VC). The AC and VC contain the same 3,371 videos from the MSR-VTT, and each video clip is annotated with two audio captioning sentences and two visual captioning sentences by different workers, respectively. So, each dataset has 6,742 clip-sentence pairs.

\subsection{Visual and Audio Representations}
\label{feas}
We use pre-trained CNN models to extract visual and audio features for training our captioning networks. 

\vspace{2mm}
\noindent\textbf{Visual Representation:} \quad 
For each visual clip $V$, we uniformly sample 40 frames and use ResNet-152~\cite{he2016deep} trained on the ImageNet~\cite{deng2009imagenet} to extract 2,048-D spatial visual feature vector for each frame. We extract a 2,048-D spatio-temporal feature vector from 16 frames in the $V$ using a 3D ResNet~\cite{hara3dresnets} trained on the Kinetics dataset~\cite{carreira2017quo}. We append the spatio-temporal feature vector to each spatial visual feature of $V$, and obtain 40 4,096-D feature vectors. In practice, we use a FC layer to reduce the dimension of these feature vectors to 512.

\vspace{2mm}
\noindent\textbf{Audio Representation:} \quad 
For each audio clip $A$, we select the first 20s audio segment to extract features for batch-wise training. If the length of the audio clip is smaller than 20s, we will append it with zeros. We use VGGish~\cite{45611} model pre-trained on the AudioSet~\cite{45857} to extract audio features $a\in\mathcal{R}^{20\times 128}$. 

\vspace{2mm}
\noindent\textbf{Word Representation:} \quad 
We obtain the vocabulary with 16,860 words from the MSR-VTT dataset by ignoring words with frequency less than two. The dimension of the word embedding is set as 512. 

\subsection{Evaluation Metrics}
\label{metric}
We use four commonly used automatic evaluation metrics: Bilingual  Evaluation Understudy (BLEU)~\cite{papineni2002bleu}, Metric for Evaluation of Translation with Explicit Ordering (METEOR)~\cite{banerjee2005meteor}, ROUGE-L~\cite{lin2004rouge}, and Consensus based Image Description Evaluation (CIDEr)~\cite{vedantam2015cider} to measure similarity between ground truth and automatic video description results. We adopt the publicly available evaluation code from Microsoft COCO image captioning evaluation server~\cite{chen2015microsoft}. 

\subsection{Implementation Details}
\label{ID}
The hidden state sizes of the visual LSTM and audio LSTM are 512 and 128, respectively. We use 10 residual blocks in the audio-text MM-CNN and visual-text MM-CNN. We set training batch size as 96 and exploit Adam~\cite{kingma2014adam} to optimize the network. The learning rate is initially set as $5\times10^{-4}$ and then reduced by a factor of 0.5 for every 10 epochs. The maximum epoch number is 50. We implement our algorithms using Pytorch~\cite{paszke2017automatic}.

\subsection{Comparison with State-of-the-art Methods}
\label{sota}
\setlength{\tabcolsep}{2pt}
\begin{table}[t!]
\begin{center}
\caption{Performances of the proposed model and other state-of-the-art methods on MSR-VTT dataset~\cite{xu2016msr}. All values are reported as a percentage (\%).}
\label{SOTA}
\begin{tabular}{l | c  c c c}
\toprule
 Models& BLEU-4 &METEOR    &ROUGE-L  &CIDEr \\
\midrule
S2VT~\cite{venugopalan2015sequence}          &31.4 &25.7 &55.9 &35.2\\
Alto~\cite{shetty2016frame}          &39.8 &26.9 &59.8 &45.7\\
VideoLab~\cite{ramanishka2016multimodal}      &39.1 &27.7 &60.6 &44.4\\
v2t-navigator~\cite{jin2016describing} &40.8 &28.2 &61.1 &44.8\\
TDDF~\cite{Zheng_2017_CVPR}          &37.3 &27.8 &59.2 &43.8\\
MA-LSTM~\cite{xu2017learning}       &36.5 &26.5 &59.8 &41.0\\
DenseVidCap~\cite{shen2017weakly}   &\textcolor{blue}{41.4} &\textcolor{blue}{28.3} &61.1 &\textcolor{red}{48.9}\\
M3~\cite{wang2018m3}                                &38.1 &26.6 &- &- \\
RecNet~\cite{wang2018reconstruction} &39.1 &26.6 &59.3 &42.7\\  
PickNet~\cite{chen2018less}       &38.9 &27.2 &59.5 &42.1\\
HRL~\cite{wang2018video}           &41.3 &\textcolor{red}{28.7} &\textcolor{red}{61.7} &\textcolor{blue}{48.0}\\
\midrule
MM-CNN           &\textcolor{red}{42.2} &\textcolor{blue}{28.3} &\textcolor{blue}{61.4} &\textcolor{blue}{48.0}\\
\bottomrule
\end{tabular}
\end{center}
\vspace{-5mm}
\end{table}

We compare the proposed MM-CNN based video captioning network with state-of-the-art models: S2VT~\cite{venugopalan2015sequence}, Alto~\cite{shetty2016frame} , VideoLab~\cite{ramanishka2016multimodal} , v2t-navigator~\cite{jin2016describing}, TDDF~\cite{Zheng_2017_CVPR} , MA-LSTM~\cite{xu2017learning} , DenseVidCap~\cite{shen2017weakly}, M3~\cite{wang2018m3}, RecNet~\cite{wang2018reconstruction}, PickNet~\cite{chen2018less}, and HRL~\cite{wang2018video}. Note that, unlike some compared methods (\eg. \cite{shen2017weakly}), we do not adopt augmentation during the training procedure. 

Table \ref{SOTA} shows the performances of the proposed model and other state-of-the-art methods on MSR-VTT dataset. The proposed MM-CNN achieves the best performance on the BLEU-4 metric among all compared networks. Except the reinforcement learning-based method, HRL, which directly utilizes CIDEr score as the reward to optimize the network, the MM-CNN and DenseVidCap outperform the other state-of-the-art networks on METEOR and the MM-CNN achieves better performance than others on the ROUGE-L metric. Our MM-CNN has the same CIDEr results with the HRL and is slightly worse than the DenseVidCap, which utilizes data augmentation during training and exploits additional lexical labels. 

Overall, the proposed audio-visual video captioning network can achieve comparable performance with the current state-of-the-art models without bells and whistles, which demonstrates the potential of the MM-CNN-based video captioning architecture. 

\begin{figure*}
\centering
\includegraphics[width=\textwidth]{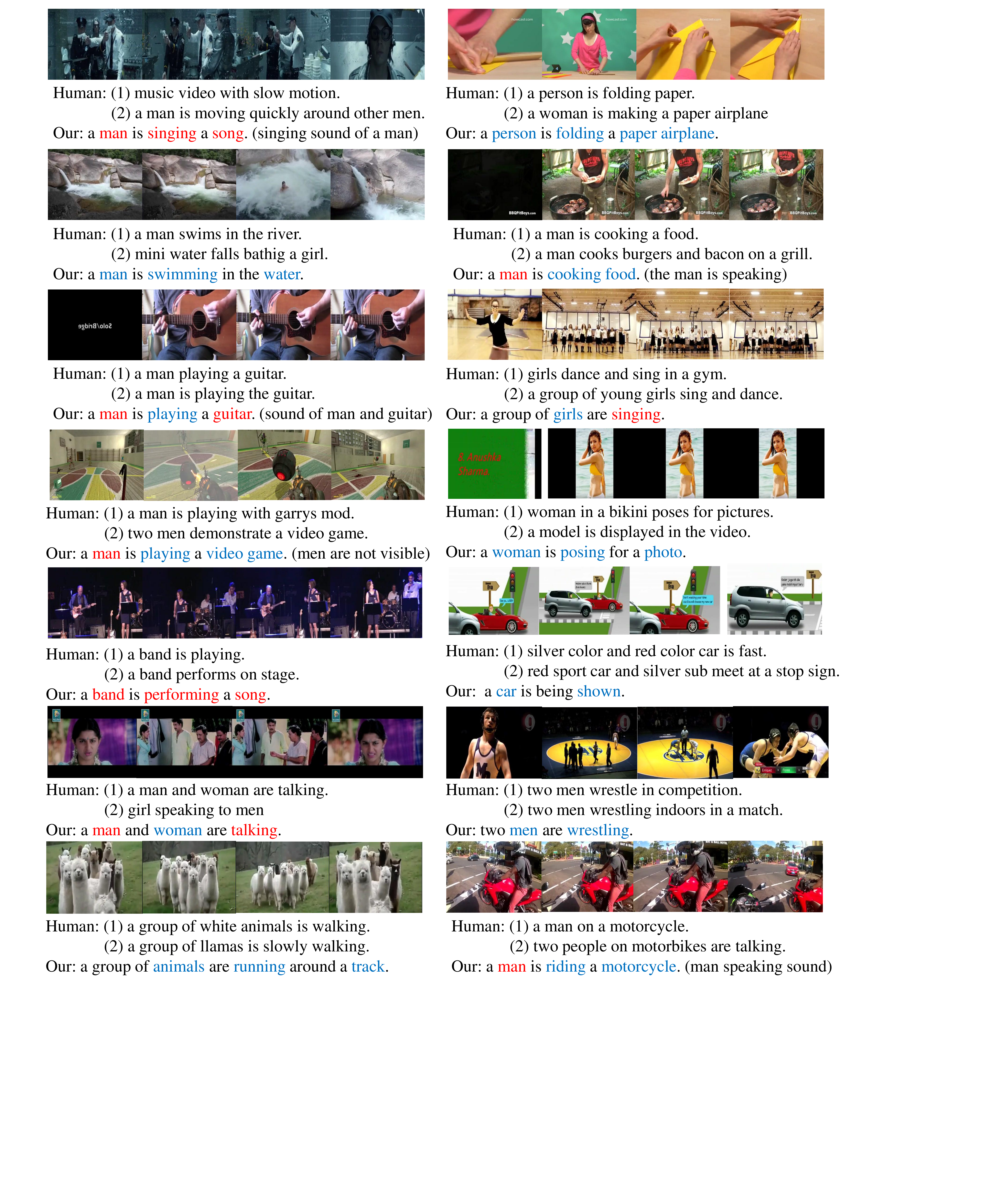}
\caption{Audio-visual video captioning results with modality selection visualizations. Here, audio activated words and visual activated words are highlighted with \textbf{red} and \textbf{blue} texts, respectively. We can find that the proposed audio-visual video captioning network can predict accurate and meaningful sentences. The activation visualization results also validate the effectiveness of the modality-aware aggregation with activation energy as the measurement.}
\label{cap_res}
\end{figure*}
\subsection{Interpretability on Modality Selection}
\label{res}

Figure~\ref{cap_res} illustrates audio-visual video captioning results with modality selection visualizations to validate the interpretability of the proposed model. 

We see that our modality selection module is capable of handling various video events, such as swimming, singing, paper cutting, and wrestling. For the first example in the first row, the generated sentence is \textit{a man is singing a song}, which is corresponding to audio content of the video. When predicting words \textit{man}, \textit{singing}, and \textit{song}, auditory modality is activated. For the second example in the first row, visual modality is activated when predicting the words \textit{person}, \textit{folding}, \textit{paper}, and \textit{airplane}. Our network generates an interesting sentence for the \textit{video game} example (first one in the 4th row), sound sources (two men) are not visible, but our network predicts the word \textit{man} by activating auditory modality. 

These modality selection results demonstrate the effectiveness and interpretability of the proposed modality-aware aggregation module with the activation energy as the criteria. In addition, the captioning results also validate the efficiency of our MM-CNN-based audio-visual video captioning network.

\setlength{\tabcolsep}{2pt}
\begin{table}[t!]
\begin{center}
\caption{Captioning results on the Audio Captioning and Visual Captioning datasets. A2A and V2A models predict audio captions with audio features and visual features as inputs respectively. Similarly, V2V and A2V predict visual captions with visual features and audio features as inputs respectively.}
\label{individual}
\begin{tabular}{l | c| c  c c c}
\toprule
 Models& Dataset &BLEU-4 &METEOR    &ROUGE-L  &CIDEr \\
\midrule
A2A & AC &6.1 &10.0 &29.1 &14.4\\
V2A & AC &6.0 &10.2 &30.1 &16.2\\
\midrule
V2V & VC &7.2 &12.3 &31.6 &21.3\\
A2V & VC &6.8 &10.7 &30.9 &16.8\\
\bottomrule
\end{tabular}
\end{center}
\vspace{-5mm}
\end{table}
\subsection{Audio, Visual, and Cross-Modal Captioning}
\label{cmc}
Although interactions between auditory and visual modalities have been studied on audio-visual representation learning~\cite{owens2016ambient,aytar2016soundnet} and recognition~\cite{Tian_2018_ECCV}, they have never been explored on high-level video understanding tasks, like captioning in this task. To explore the interaction between the two modalities and study whether auditory modality can infer high-level semantics (language descriptions) of visual modality, and vice versa, we conduct extensive experiments using the newly-collected VC and AC datasets with the audio, visual, and cross-modal captioning as a testbed. Unlike the audio-visual video captioning network, which contains audio and visual branches, we only select an audio or a visual branch for these tasks depending on the input modality.

Four tasks, A2A, V2A, V2V, and A2V, are investigated. Here, we define the A2A task as given an audio clip to generate an audio description; the V2A task as given a visual clip to generate a visual description; the V2V task as given a visual clip to generate a visual description; the A2V task as given an audio clip to generate a visual description. The A2A and V2V are audio captioning and visual captioning tasks, and the V2A and the A2V are cross-modal captioning tasks. 

Table \ref{individual} shows captioning performances\footnote{The scores are much smaller than the audio-visual video captioning results on the full MSR-VTT dataset due to limited sizes of the AC and VC datasets.} of the above four tasks. As expected, V2V is better than A2V, which demonstrates that visual features are more powerful than audio features for the visual captioning task. A2V achieves slightly worse performance than the V2A, which validates the potential of inferring high-level semantics of the visual modality from the auditory modality. An interesting observation is that V2A is even slightly better than A2A. We note that AMT workers only listened to the audio channel to write language sentences while annotating the AC dataset, but audio features fail to achieve better audio captioning performance, which demonstrates that visual information is more useful for the task. The reasons are three-fold. (1) Visual content indicates sound (\eg, if a woman on a stage, we can imagine that there may be speech sound or a singing sound of the woman). (2) Audio contents may be noisy in some videos (\eg, background sound: music, outdoor natural sound noise, and crowd noise. For these cases, the sound source may be visible in the visual contents). (3) Humans may learn the event in an audio clip from speech, but the current model has no capability for speech-to-text translation. For example, a woman teaches cooking in the audio clip. It is difficult to infer the event directly from the input audio clip by our model. However, we may easily observe the lip movements of the woman and the cooking event. 

Based on the above the observations, we know that there is a strong correlation between the auditory and the visual modalities, and two modalities have capacities to infer each others' high-level semantics.  

\setlength{\tabcolsep}{2pt}
\begin{table}[t!]
\begin{center}
\caption{Performances of the proposed model with different numbers of residual blocks on MSR-VTT dataset~\cite{xu2016msr}. All values are reported as a percentage (\%).}
\label{depth}
\begin{tabular}{l | c  c c c}
\toprule
 Models& BLEU-4 &METEOR    &ROUGE-L  &CIDEr \\
\midrule
5B          &40.9 &\textbf{28.5} &60.5 &46.2\\
10B  &\textbf{42.2} &28.3 &\textbf{61.4} &\textbf{48.0}\\
15B     &40.2 &27.5 &60.0 &44.0\\
\bottomrule
\end{tabular}
\end{center}
\vspace{-5mm}
\end{table}

\subsection{Effect of Numbers of the Residual Block}
\label{block}
Table \ref{depth} presents the performances of the proposed MM-CNN with different numbers, e.g., 5, 10, and 15, of the residual block. We can find that the network with 10 blocks achieves overall better results than with 5 or 15 blocks. 15B is worse than 5B, which demonstrates that the even deep models fail to further improve performance due to the limitation of the size of the training dataset. We may alleviate the issue with further exploring some data augmentation techniques, like randomly sampling frames as in~\cite{shen2017weakly}.

\section{Conclusion}
\label{conclude}

In this work, we propose a novel multimodal convolutional neural network-based audio-visual video captioning framework equipping with a modality-aware feature aggregation module, which makes the framework have strong interpretability on modality selection during word generation. We also find that there is a strong correlation between the auditory and the visual modalities, and two modalities have capacities to infer each others’ high-level semantics. Experimental comparison with other video captioning networks validates the effectiveness of the proposed multimodal convolutional neural network-based network. In the future, we would like to extend the multimodal convolutional neural network to address other multimodal tasks, such as video question answering and video-to-text retrieval. 

\section{Acknowledgement}
This work was supported in part by NSF IIS-1741472, IIS-1813709, and IIS-1659250. Any opinions, findings, and conclusions or recommendations expressed in this material are those of the authors and do not necessarily reflect the views of the NSF.

{\small
\bibliographystyle{ieee}
\bibliography{egbib}
}

\end{document}